\documentclass[11pt,a4paper]{article}
\usepackage[hyperref]{acl2021}
\usepackage{times}
\usepackage{latexsym}

\usepackage{microtype}

\aclfinalcopy 

\usepackage[switch]{lineno}
\usepackage{booktabs}
\usepackage{amsmath}
\usepackage{amssymb}
\usepackage{enumitem}
\usepackage{adjustbox}
\usepackage[labelformat=simple]{subcaption}
\usepackage{makecell}
\usepackage{multirow}
\usepackage{mathtools}
\usepackage{placeins}

\usepackage{color, colortbl}
\definecolor{LightCyan}{rgb}{0.88,1,1}
\definecolor{LightRed}{rgb}{1.0, 0.91, 0.91}
\definecolor{LightGray}{rgb}{0.88,0.88,0.88}
\definecolor{VeryLightGray}{rgb}{0.93,0.93,0.93}

\title{
WeaQA: Weak Supervision via Captions for Visual Question Answering
}

\author{
Pratyay Banerjee \quad Tejas Gokhale \quad Yezhou Yang \quad Chitta Baral  \\
Arizona State University \\
\texttt{pbanerj6, tgokhale, yz.yang, chitta}@asu.edu               \\
}

\date{}

\begin{document}
\maketitle

\begin{abstract}
Methodologies for training visual question answering (VQA) models assume the availability of datasets with human-annotated \textit{Image-Question-Answer} (I-Q-A) triplets.
This has led to heavy reliance on datasets and a lack of generalization to new types of questions and scenes.
Linguistic priors along with biases and errors due to annotator subjectivity have been shown to percolate into VQA models trained on such samples.
We study whether models can be trained without any human-annotated Q-A pairs, but only with images and their associated textual descriptions or captions.
We present a method to train models with synthetic Q-A pairs generated procedurally from captions.
Additionally, we demonstrate the efficacy of spatial-pyramid image patches as a simple but effective alternative to dense and costly object bounding box annotations used in existing VQA models.
Our experiments on three VQA benchmarks demonstrate the efficacy of this weakly-supervised approach, especially on the VQA-CP challenge, which tests performance under changing linguistic priors.
\end{abstract}

\section{Introduction}
Since Visual Question Answering (VQA) was first proposed as a Turing test~\cite{malinowski2014towards}, several human-annotated datasets~\cite{mogadala2019trends}
have been used to train and evaluate VQA models.
Unfortunately, heavy reliance on these datasets for training has the unwanted side-effects of bias towards answer styles, question-types~\cite{chao2018cross}, and spurious correlations with language priors~\cite{agrawal2018don}.
Similar findings have been reported for natural language tasks~\cite{gururangan2018annotation,niven2019probing,kaushik2019learning}.
Evaluating VQA models on test-sets that are very similar to training sets is deceptive and inadequate and not an accurate measure of robustness.

To address this, one line of work has focused on balancing, de-biasing, and diversifying samples~\cite{goyal2017making,zhang2016yin}.
However, crowd-sourcing ``unbiased" labels is difficult and costly; it requires a well-designed annotation interface and a large-scale annotation effort with dedicated and able annotators~\cite{sakaguchi2020winogrande}.
The alternative (that this paper aligns itself with) is to avoid the use of explicit human annotations and instead to train models in an unsupervised manner by synthesizing training data.
These techniques, coined \textit{unsupervised}\footnotemark[1],
come with many advantages -- human bias and subjectivity are reduced; the techniques are largely domain-agnostic and can be transferred from one language to another (low resource languages) or from one visual domain to another.
For instance, template-based Q-A generation developed for synthetic blocks-world images in CLEVR~\cite{johnson2017clevr} can also be used to generate Q-A pairs for natural complex scenes in GQA~\cite{hudson2019gqa} or the referring-expressions task~\cite{liu2019clevr}.

In this work, we train VQA models without using human-annotated Q-A pairs.
Instead, we rely on weak supervision from image-captioning datasets, which provide multi-perspective, concise, and less subjective descriptions of visible objects in an image.
We procedurally generate Q-A pairs from these captions and train models using this synthetic data, and \textit{only evaluate} them on established human-annotated VQA benchmarks.

\textbf{Why Captions?} Image captioning, like VQA, has been a central area of vision-and-language research.
Datasets such as MS-COCO~\cite{lin2014microsoft,chen2015microsoft} contain captions that describe objects and actions in images of everyday scenes.
During the construction of MS-COCO, human captioners were instructed to refrain from describing past and future events or ``what a person might say".
On the other hand, annotators of VQA~\cite{antol2015vqa} were instructed to ask questions that \textit{``a smart robot cannot answer, but a human can''} and ``interesting'' questions that may require ``commonsense''.
Different sets of annotators provided answers to these questions and were allowed to speculate or even guess an answer that \textit{most people would agree on}.
It has also been shown that multiple answers may exist for questions in common VQA datasets~\cite{bhattacharya2019does}.

\begin{figure*}[t]
    \centering
    \includegraphics[width=\linewidth]{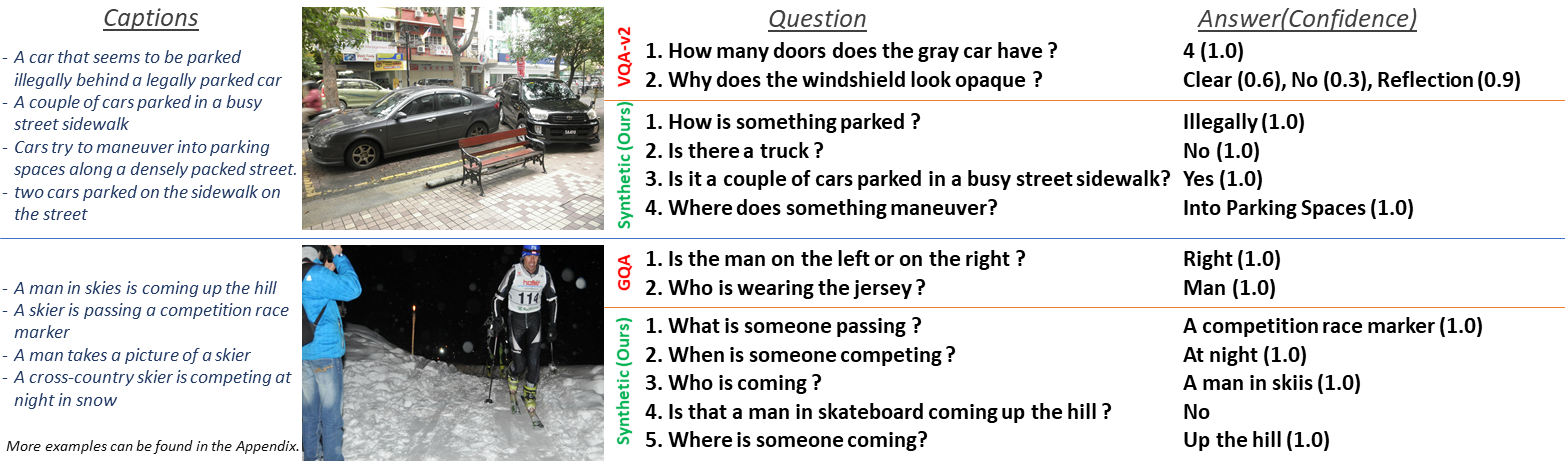}
    \caption{Examples of images and human-annotated Q-A pairs from VQA and GQA and our synthetic Q-A pairs.}
    \label{fig:data}
\end{figure*}

In Figure~\ref{fig:data}, the first VQA-v2 question asks how many doors the car has.
Although commonsense (and linguistic priors) would suggest that ``Most cars have \textit{four} doors", only two doors can be seen in the image.  What should the model predict, \textit{two} or \textit{four}? 
The second question is subjective and has multiple contradicting answers from different annotators (where one should draw the line between opaque, transparent, or reflective is not very clear).
Similarly, the first GQA question is ambiguous and could refer to either the skier or the photographer.

Thus the very nature of the data-collection procedure and instructions for VQA brings in human subjectivity and linguistic bias as compared to caption annotations, which are designed to be simple, precise, and non-speculative. 
Motivated by this, we study the benefits of using captions to synthesize Q-A pairs, using three types of methods:
\begin{enumerate}[nosep,noitemsep]
    \item template-based methods similar to~\cite{ren2015exploring,gokhale2020vqa},
    \item paraphrasing and back-translation~\cite{sennrich2016improving} which provide linguistic variation,
    \item synthesis of questions about image semantics using the QA-SRL~\cite{he2015question} approach.
\end{enumerate} 
Since our Q-A pairs are created synthetically, there does exist a domain shift as well as label (answer) shift from evaluation datasets such as VQA-v2 and GQA as shown in Figure~\ref{fig:data}, thus posing challenges to this weakly-supervised method.

We evaluate two models, UpDown~\cite{Anderson_2018_CVPR} and a transformer-encoder~\cite{vaswani2017attention} based model pre-trained on synthetic Q-A pairs and image-caption matching task.
To remove the dependence on object bounding-boxes and labels needed to extract object features, 
we propose spatial pyramids of image patches as a simple and effective alternative.

To the best of our knowledge, this is the first work on the unsupervised\footnote{adhering to the usage of this term in~\citet{lewis2019unsupervised}.} visual question answering, with the following contributions:
\begin{itemize}[noitemsep,nosep]
    \item We introduce a framework for synthesizing \textit{(Question, Answer)} pairs from captions.
    \item Since synthetic samples (unlike popular benchmarks) include multi-word answer phrases, we propose a sub-phrase weighted-answer loss to mitigate bias towards such multi-word answers.
    \item We propose pre-training tasks that use spatial pyramids of image-patches instead of object bounding-boxes, further removing the dependence on human annotations.
    \item Extensive experiments and analyses under zero-shot transfer and fully-supervised settings on VQA-v2, VQA-CP, and GQA show our model's efficacy and establish a strong baseline for future work on unsupervised visual question answering.
\end{itemize}

\section{Related Work}
\begin{figure}[t]
    \centering
    \includegraphics[width=\linewidth]{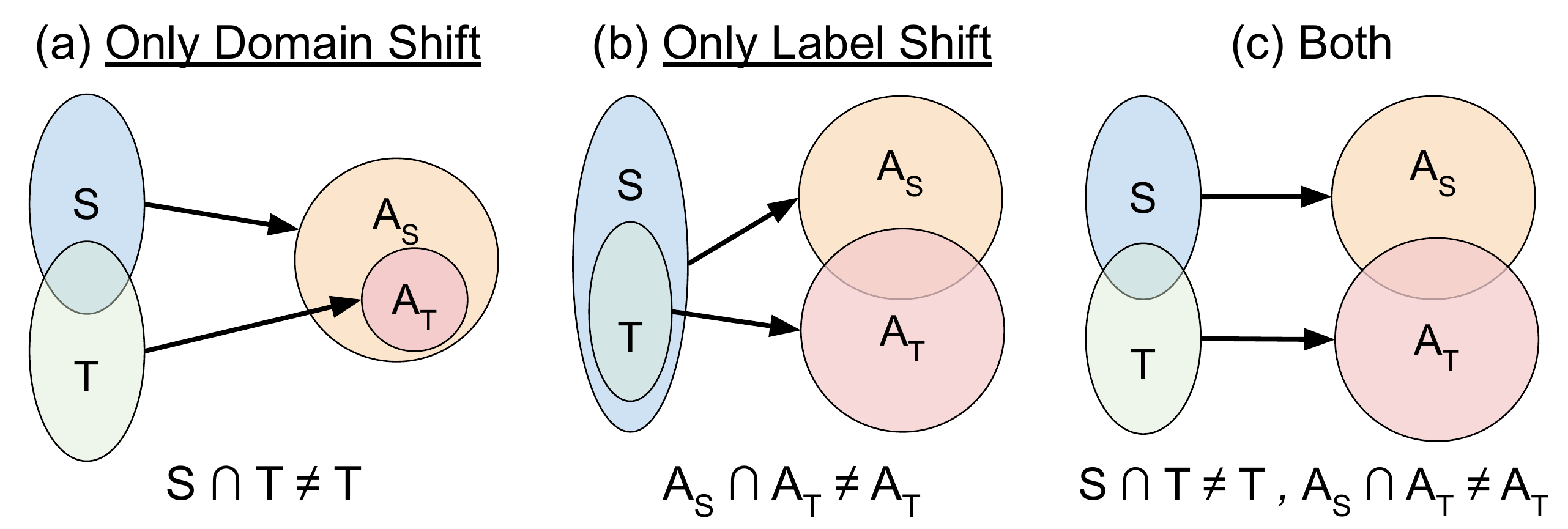}
    \caption{Aspects of generalization in VQA.
    }
    \label{fig:generalization_types}
\end{figure}

\textbf{Robustness in VQA}
can be defined as shown in Figure~\ref{fig:generalization_types} under two situations: domain shift and label shift.
Under domain shift, generalization to a new input domain (such as different styles of questions or novel scenes) is desired, characterized by $S\cap T \neq T$ where $S$ and $T$ denote the train and test input domains.
Under label shift, generalization to novel answers is desired (predicting answers not seen during training), characterized by $A_S\cap A_T \neq A_T$, where $A_S$ and $A_T$ are the set of answers seen during training and test-time.

Performance under \textbf{domain shift} has been evaluated for new domains of test questions with unseen words and objects~\cite{teney2016zero,ramakrishnan2017empirical}, 
novel compositions~\cite{johnson2017clevr,agrawal2017c}, 
logical connectives~\cite{gokhale2020vqa}, as well as questions that are
implied~\cite{ribeiro-etal-2019-red}, entailed~\cite{ray2019sunny} or sub-questions~\cite{selvaraju2020squinting}; or for datasets with varying linguistic styles~\cite{chao2018cross,xu2019open,shrestha2019answer}
and different reasoning capabilities~\cite{kafle2017analysis}.

\textbf{Label shift} or Prior Probability Shift~\cite{storkey2009training} has been implicitly explored in VQA-CP~\cite{agrawal2018don}, where the conditional probabilities of answers given the question type deviate at test-time.
\citet{teney2020value} have identified several pitfalls associated with the models and evaluation criteria for VQA-CP.

\textbf{Unsupervised Extractive QA} 
in which aligned {\it(context, question, answer)} triplets are not available, has been studied~\cite{lewis-etal-2019-unsupervised,banerjee-baral-2020-self,rennie-etal-2020-unsupervised,fabbri-etal-2020-template,li-etal-2020-harvesting, banerjee-etal-2021-self} by training models on procedurally generated Q-A pairs.
Captions have been used to generate Q-A pairs for logical understanding~\cite{gokhale2020vqa} and commonsense video understanding~\cite{fang-etal-2020-video2commonsense}.
\citet{li2018visual,krishna2019information} have explored Visual Question Generation from an input image and answer.

\textbf{Weak supervision} is an active area of research; for instance in action/object localization~\cite{song2014learning,zhou2016learning} and semantic segmentation~\cite{khoreva2017simple,zhang2017ppr} without pixel-level annotations, but only class labels.
There is also interest growing in leveraging natural language captions or textual queries as weak supervision for visual grounding tasks~\cite{anne2017localizing,mithun2019weakly,fang2020weak}.

\textbf{Visual Feature Extractors} such as 
VGG~\cite{Simonyan2015VeryDC} and  
ResNet~\cite{he2016deep} have been widely used for many computer vision tasks.
Object-based features such as RCNN~\cite{girshick2014rich} and Faster-RCNN~\cite{ren2015faster}
have become the standard for V\& L tasks~\cite{Anderson_2018_CVPR}.

\section{Framework for Synthesizing Q-A Pairs}
\label{sec:3}
\begin{figure*}[t]
    \begin{subfigure}{.22\textwidth}
        \includegraphics[width=\linewidth,height=27mm]{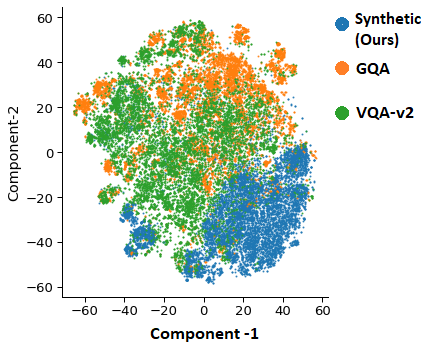}
    \end{subfigure} ~
    \begin{subfigure}{.76\textwidth}
        \resizebox{\linewidth}{!}{
            \centering
            \begin{tabular}{@{}>{\raggedright} p{35mm}cccccc@{}}
                \toprule
                 & \thead{Template-based} & \thead{Paraphrase \& \\ Back-translate} & \thead{QA-SRL} & \thead{VQA-v2} & \thead{GQA} & \thead{VQA-CP} \\ 
                \midrule
                \# of Questions         & 600K & 400K & 2.5M & 438K / 214K & 943K / 132K & 245K / 220K \\
                \# of Answers           & 5K & 5K & 90K & 3.5K & 1878 & 3.5K \\
                Mean Question Length    & 7.9 & 8.1 & 4.8 & 6.4 & 10.6 & 6.4 \\
                Mean Answer Length      & 1.4 & 1.4 & 6.3 & 1.1 & 1.3 & 1.1 \\ 
                Image Source            & COCO & COCO & COCO & COCO & COCO,VG,Flickr & COCO \\
                Image Counts            & 120K &  120K &  120K &  120K & 113K & 120K \\
                \bottomrule
            \end{tabular}
        }
    \end{subfigure}
    \caption{Discrepancy between VQA-v2, GQA, and synthetic samples. Left: t-SNE plot of question 
    embeddings. Right: Dataset statistics for our generated Q-A pairs with Train/Val.\ splits for benchmark datasets.
    }
    \label{fig:data_tsne_stats}    
\end{figure*}

\paragraph{Problem Statement:}
Consider a dataset containing images and associated captions 
as shown in Figure~\ref{fig:data}.
Our work deals with learning VQA using these image-caption data, without any labeled Q-A pairs, and answer questions about unseen images.


\subsection{Question Generation}
Several studies~\cite{du2017learning,lewis2019unsupervised} have been dedicated to the complex domain of question generation.
We approach it conservatively, using template-based methods and semantic role labeling, with paraphrasing and back-translation for improving the linguistic diversity of template-based questions.
We begin by extracting object words from the caption by using simple heuristics such as extracting noun-phrases and using numerical quantifiers in the caption as soft approximations of objects' cardinality.  
If object-words are available explicitly, we used them as is.
Questions are categorized based on answer types; \textit{Yes-No}, \textit{Number}, \textit{Color}, \textit{Location}, \textit{Object}, and \textit{Phrases}.

\paragraph{Template-based:}
To create \textit{Yes-No} questions, 
modal verbs are removed from the caption, and a randomly chosen question prefix such as \textit{``is there", ``is this"} is attached.
For instance, the caption ``A man is wearing a hat and sitting" is converted to ``\textit{Is there a} man wearing a hat and sitting", with the answer ``Yes".
To create the corresponding question with the answer ``No", we use either negation or replace the object-word with an adversarial word or antonym, thus obtaining ``Is there a dog wearing a hat and sitting" for which the answer is ``No". 
An adversarial word refers to an object absent in the image but similar to objects in the image.
To compute similarity, we use Glove~\shortcite{pennington2014glove} word-vectors.

For \textit{Object}, \textit{Number}, \textit{Location}, and \textit{Color} questions, we follow a procedure similar to~\citet{ren2015exploring}. 
To create ``\textit{what}" questions for the \textit{Object} type, we extract objects and noun phrases from captions as potential answers and replace them with \textit{what}.
The question is rephrased by splitting long sentences into shorter ones and converting indefinite determiners to definite.
A similar procedure is used for \textit{Number} questions; numeric quantifiers of noun phrases are extracted and replaced by ``how many'' and ``what is the count'' to form the question.
\textit{Color} questions are generated by locating the color adjective and the corresponding noun phrase and replacing them in a templated question: ``What is the {color} of the {object}?". 
\textit{Location} questions are similar to \textit{Object} questions, but we extract phrases with ``in'', ``within'' to extract locations, with places, scenes, and containers as answers.
        
\paragraph{Semantic Role Labeling:}
QA-SRL~\cite{he2015question} was proposed as a paradigm to use natural language to annotate data by using Q-A pairs to specify textual arguments and their roles.
Consider the caption \textit{``A girl in a red shirt holding an apple sitting in an empty open field"}.
Using QA-SRL with B-I-O span detection and sequence-to-sequence models~\cite{fitzgerald2018large}, for the \textit{``when'', ``what", ``where'', and ``who''} questions, we obtain Q-A pairs belonging to the \textit{Phrases} category such as:

\noindent
\resizebox{\linewidth}{!}{
    \fbox{
        \parbox{\linewidth}{
            \small \it \centering
            (what is someone holding?,~~~~an apple)\\ 
            (who is sitting?,~~~~girl in a red shirt holding an apple)\\
            (where is someone sitting?,~~~~an empty open field)
        }
    }
}

These examples illustrate that QA-SRL questions are short and use generic descriptors such as \textit{something} and \textit{someone} instead of elaborate references, while the expected answer phrases are longer and descriptive.
Thus to answer these, better semantic image understanding is required.
        
\paragraph{Paraphrasing and Back-Translation (P{\&}B):} 
We apply two natural language data augmentation techniques, paraphrasing, and back-translation to increase the linguistic variation in the questions. 
To paraphrase questions, we train a T5~\cite{raffel2019exploring} text generation model on the Quora Question Pairs Corpus~\shortcite{WinNT}.
For back-translation, we train another T5 text generation model on the Opus corpus~\shortcite{TIEDEMANN12.463}, translate the question to an intermediate language (Fran\c cais, Deutsche, or Espa\~nol), and translate the question back to English.  
For example:

\noindent
\resizebox{\linewidth}{!}{
    \fbox{
        \parbox{\linewidth}{
            \small
            \centering
            \textit{Is the girl who is to the left of the sailboats wearing a backpack?}\\
            $\Big\downarrow{\text{Espa\~nol}}$\\
            \textit{La chica que está a la izquierda de los veleros lleva mochila?}\\
            $\Big\downarrow{\text{English}}$\\
            \textit{Does the girl to the left of the sailboats carry a backpack?}
        }
    }
}
        
\subsection{Domain Shift w.r.t.\ VQA-v2 and GQA}
Compared to current VQA benchmarks (which typically contain one-word answers), answers to QA-SRL questions are more descriptive and contain adjectives, adverbs, determiners, and quantifiers, as seen in Figure \ref{fig:data}. 
On the other hand, synthetic questions have less descriptive subjects due to the use of pronouns.
Our synthetic data contains $90k$ unique answer phrases, compared to $3.2k$ in VQA and 3k in GQA. 
Around 200 answers from VQA are not present in our answer phrases, such as time (11:00) and proper nouns (LA Clippers), both of which are not present in caption descriptions.

Moreover, our training data contains Q-A pair such as {\it(``Where is the man standing?, ``to the left of the table'')}, generated by QA-SRL with long phrases as answers. However, the test set contains questions such as {\it(``Which side of the car is the tree?'', ``left'')}, which expects only {\it``left''} as the answer. So although the word {\it``left''} is seen as a sub-phrase of our training answers, it is not explicitly seen as an only correct answer.

Some of our synthetic template-based questions about counting and object presence are similar in style to those in VQA and GQA.
However, QA-SRL questions require a semantic understanding of the actions depicted in the image, which are rare in VQA and GQA. 
We quantify this by plotting the t-SNE components of document vector embeddings of the questions from VQA, GQA, and our synthetic data, in Figure \ref{fig:data_tsne_stats}, and observe 
that our synthetic questions are a distinct cluster, while VQA and GQA overlap with each other.
As such, a linguistic domain shift exists between these synthetic source questions and human-annotated target questions.
In this paper, we address the challenge of learning VQA on a synthetically generated dataset and evaluating models on conventional benchmarks which have questions and answers that deviate linguistically from synthetic training samples.


\begin{figure*}[t]
    \centering
    \includegraphics[height=35mm,width=\linewidth]{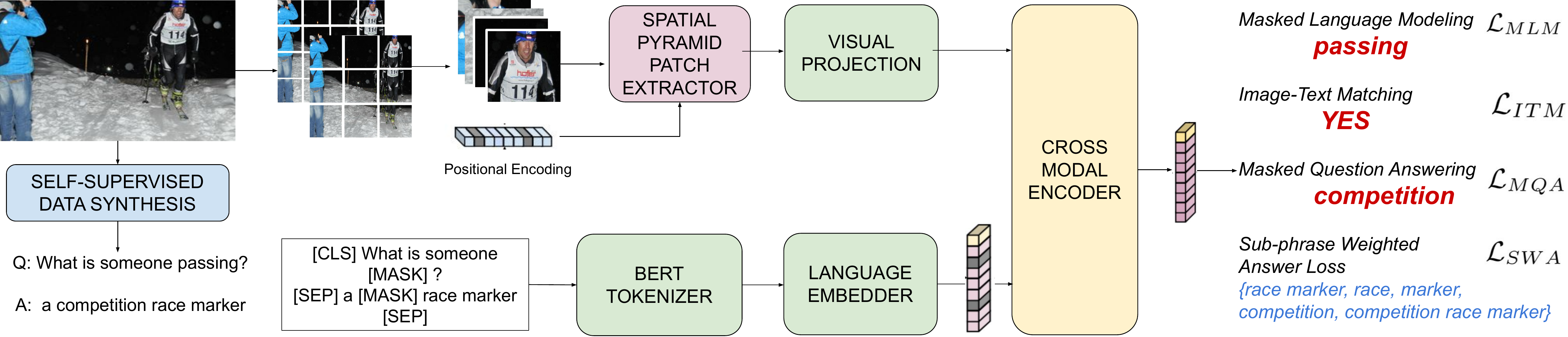}
    \caption{Our model architecture makes the use of spatial pyramids of image patches as inputs to the Encoder, which is trained for three pre-training tasks as shown.}
    \label{fig:model}
\end{figure*}        

\section{Method}
Recently, multiple deep transformer-based architectures have been proposed~\cite{tan2019lxmert,lu2019vilbert,chen2019uniter}, that are pretrained on a combination of multiple VQA and image captioning datasets such as Conceptual Captions~\cite{sharma2018conceptual}, SBU Captions~\cite{ordonez2011im2text}, Visual Genome~\cite{krishna2017visual}, and MSCOCO~\cite{lin2014microsoft}. 
These models are resource intensive as they are trained on a huge collection of data with $~3$ million images.
We train our models only on MS-COCO captions and images (${\sim}204$k), without access to any human-authored Q-A pairs or object bounding boxes.

\subsection{Spatial Pyramid Patches}
``Bottom-Up" object features~\cite{Anderson_2018_CVPR} extracted from Faster R-CNN~\cite{ren2015faster} have become the de-facto features used in state-of-the-art VQA models.
These VQA models thus only use features of detected objects as input, and ignore the rest of the image.
Although object features are discriminative, dense annotations are required for training and additional large deep networks for extraction.
Object detection can be imperfect for small and rare objects~\cite{wang2019meta}; for instance
if an object detection model detects only four out of six bananas in an image, features of the other two bananas will not be used by VQA models.
This creates a performance bottle-neck for questions about counting or rare objects.

We take a step back and postulate that the use of features of the entire image in context could reduce this bottleneck.
Image features extracted from a ResNet~\cite{he2016deep} trained for the ImageNet~\cite{russakovsky2015imagenet} classification task, which is widely used for computer vision tasks, have been previously used for VQA models~\cite{goyal2017making}.
Unfortunately, since ImageNet contains iconic (single-object) images, using these features for non-iconic VQA images is restrictive since many questions refer to multiple objects and backgrounds in the image. 
Inspired by Spatial Pyramid Matching~\cite{lazebnik2006beyond} for image classification, we propose \textit{spatial pyramid patch features} to represent the input VQA image into a sequence of features at different scales.

We divide each image $I$ into a set of image patches $\{ I_{k_1},\dots, I_{k_n}\}$, each $I_{k_i}$ being a $k_i\times k_i$ grid of patches, and extract ResNet features for each patch.
Larger patches encode global features and relations, while smaller patches encode local and low-level features.

\paragraph{Encoder:}
Our Encoder model is similar to the UNITER single-stream transformer, where the sequence of word tokens $w = \{w_1,...,w_T\}$ and the sequence of image patch features $v = \{v_1,...,v_K\}$ are taken as input. 
We tokenize the text using a WordPieces~\cite{wu2016google} tokenizer similar to BERT~\cite{devlin2018bert}, and embed the text tokens through a text-embedder~\cite{sanh2019distilbert}.
The visual features are projected to a shared embedding space using a fully-connected layer.
A projected visual position encoding, indicating the patch region (top-right, bottom-left) is added to the visual features. 
We concatenate both sequences of features and feed them 
to $L$ cross-modality attention layers. 
Parameters between the cross-modality attention layers are shared to reduce parameter count and increase training stability~\cite{lan2019albert},
and a residual connection and layer normalization is added after cross-modal attention layer similar to~\citet{vaswani2017attention}.
    
\subsection{Pre-training Tasks and Loss Functions}
We train the Encoder model using three pre-training tasks: Masked Language Modeling, Masked Question Answering, and Image-Text Matching.

\paragraph{Masked Language Modeling (MLM):}
We randomly mask 15\% of the word tokens from the caption and ask the model to predict them.
For the caption ``There is a man wearing a hat'', the model gets the input ``There is \texttt{[MASK]} wearing a hat''.
Without the image, there can be multiple plausible choices for the \texttt{[MASK]} token, such as ``woman'', ``man'', ``girl'', but given the image the model should predict ``man''.
This task has been shown to effectively learn cross-modal features~\shortcite{tan2019lxmert}.

\paragraph{Masked Question Answering (MQA):}
In this task, the answer tokens are masked, and the model is trained to predict the answer tokens.
For example in Figure \ref{fig:data}, for the input `` When is someone competing? \texttt{[MASK]} \texttt{[MASK]}'', the model should predict, ``at night''.
To answer such questions, the model needs to interpret the image.

\paragraph{Image-Text Matching (ITM):} 
We use the five captions provided by MS-COCO as positive samples for each image.
To obtain negative samples, we randomly sample captions from other images that contain a different set of objects.
We train the model on a binary classification task (matching / not matching) for each image-caption pair.

For VQA and ITM, we use the final layer representation $z^{\textit{\texttt{[CLS]}}}$ of \texttt{[CLS]} token
, followed by a feed-forward and softmax layer.
For MLM and MQA we feed corresponding token representations to a different feed-forward layer. 
We train the model using cross-entropy loss for all three tasks.

\paragraph{Sub-phrase Weighted Answer Loss:}
As observed before, the questions generated in QA-SRL have long answer phrases. 
For instance ``What is parked?'' has the answer ``two black cars''. 
We extract all possible sub-phrases that can be alternate answers, but assign them a lower weight than the complete phrase, computed as $W_{sub} = \textit{WordCount}(sub)/\textit{WordCount}(ans)$.
Thus ``two black cars'' has a  weight $1.0$, while the extracted sub-phrases and weights are: (two, 0.33), (2, 0.33), (black, 0.33), (cars, 0.33), (two cars, 0.66), (2 cars, 0.66), (black cars, 0.66), (car, 0.33).
This enforces a distribution over the probable answer space instead of a strict ``single true answer"
training. 
We train the model with this additional binary cross-entropy loss, where the model predicts a weighted distribution $y_{wa}$ over the answer vocabulary. The vocabulary is defined from the synthetic QA answer-space.
\begin{equation}
    \small
    \mathcal{L}_{SWA} =  \mathcal{L}_{BCE}(\sigma(z^{\textit{\texttt{[CLS]}}}),y_{wa}).
\end{equation}

\noindent The total loss, with scalar coefficients $\alpha,\beta \in (0, 1]$ is given by:
\begin{equation}
    \small
    \mathcal{L} = \mathcal{L}_{MLM} + \mathcal{L}_{MQA} + \alpha\cdot\mathcal{L}_{ITM} + \beta\cdot\mathcal{L}_{SWA}.
\end{equation}

\section{Experimental Setup}
\paragraph{Datasets:} 
We evaluate our methods on the three popular visual question answering benchmarks: VQA-v2, VQA-CP-v2, and GQA. 
Answering questions in VQA-v2 and VQA-CP v2 requires image and question understanding, whereas GQA further requires spatial understanding such as compositionality and relations between objects.
We evaluate our methods under \textit{zero-shot} transfer (trained only on procedurally generated samples), and \textit{fully-supervised} (where we finetune our model using the associated train annotations) settings.
We use exact-match accuracies for GQA, and use VQA-metric~\cite{agrawal2017c} for VQA.

\paragraph{Training:}
Our Encoder has 8 cross-modal layers with a hidden dimension of $768$. The weights are initialized using the standard definition as provided in the Huggingface repository~\cite{Wolf2019HuggingFacesTS}.
Our models are pre-trained for 40 epochs with a learning rate of $1\mathrm{e}{-5}$, batch size of 256, using Adam optimizer.
For finetuning, we use a learning rate of $1\mathrm{e}{-5}$ or $5\mathrm{e}{-5}$ and batch size of 32 for 10 epochs. 
We use a ResNet-50 pretrained on ImageNet to extract features from image patches with 50\% overlap, and Faster R-CNN pretrained on Visual Genome to extract object features. We evaluate both frozen and finetuned ResNet, and observe finetuning the feature extractor to perform better.
All our models are trained using 4 Nvidia V100 16 GB GPUs. All results in the fully supervised setting are reported for from-scratch trained final classification layers.

\paragraph{Baselines:}
To measure the improvements due to our proposed image patch features and SWA loss, we compare our methods to the UpDown model \citeauthor{Anderson_2018_CVPR}, which uses object bounding-box features.
For the Zero-shot transfer setting, we compare our Encoder with UpDown when trained with spatial features as well as object features.
Pre-trained transformers such as UNITER use large V\&L corpora, dense human annotations for objects and Q-A pairs
and supervised loss functions over these.
Comparisons with such models are therefore not fair in a ZSL setting; instead, we perform these comparisons in a fully-supervised (FSL) setting.

\begin{table}[t]
    \centering
    \small
    \resizebox{\linewidth}{!}{
    \begin{tabular}{@{}lcccc@{}}
        \toprule
        \textbf{Model} & \textbf{All} & \textbf{Yes-No} & \textbf{Num} & \textbf{Others} \\
         \toprule
        \rowcolor{LightCyan} SAN~\shortcite{yang2016stacked} & 25.0 & 38.4 & 11.1 & 21.7 \\
        \rowcolor{LightCyan} GVQA~\shortcite{agrawal2018don} & 31.3 & 58.0 & 13.7 & 22.1 \\
        \rowcolor{LightCyan} UpDown~\shortcite{Anderson_2018_CVPR} & 39.1 & 62.4 & 15.1 & 34.5 \\
        \rowcolor{LightCyan} AReg\shortcite{ramakrishnan2017empirical} & 42.0 & 65.5 & 15.9 & 36.6 \\
        \rowcolor{LightCyan} AdvReg~\shortcite{grand2019adversarial} & 42.3 & 59.7 & 14.8 & 40.8 \\
        \rowcolor{LightCyan} RUBi~\shortcite{RUBI} & 47.1 & 68.7 & 20.3 & 43.2 \\
        \rowcolor{LightCyan} \citet{teney2019actively} & 46.0 & 58.2 & 29.5 & 44.3 \\
        \rowcolor{LightCyan} Unshuffling~\shortcite{teney2020unshuffling} & 42.4 & 47.7 & 14.4 & 47.3 \\
        \rowcolor{LightCyan} UpDn+CE+GS~\shortcite{teney2020learning} & 46.8 & 64.5 & 15.4 & 45.9 \\
        \rowcolor{LightCyan} LXMERT~\shortcite{tan2019lxmert} & 46.2 & 42.8 & 18.9 & 55.5 \\ 
        SCR~\shortcite{wu2019self} & 48.4 & 70.4 & \cellcolor{LightRed} 10.4 & 47.3 \\
        LMH~\shortcite{clark2019don}  & 52.4 & \cellcolor{LightRed} 69.8 & 44.5 & 45.5 \\ 
        CSS~\shortcite{chen2020counterfactual}* & 58.9 & 84.4 & 49.4 & 48.2 \\
        MUTANT~\shortcite{gokhale2020mutant}* & \textbf{69.5} & \textbf{93.2} & \textbf{67.2} & \textbf{57.8} \\
        
        \midrule
        ZSL+Objects+UpDown & 40.8 & 67.4 & 28.6 & 30.2 \\
        ZSL+Patches+UpDown & 41.2 & 68.5 & 29.8 & 30.0 \\
        ZSL+Patches+Encoder & \underline{47.3} & \underline{73.4} & \underline{39.8} & \underline{35.6} \\ \bottomrule
    \end{tabular}
    }
    \caption[VQA-CP]{
        Unsupervised accuracy on VQA-CP-v2 test set.
        All baselines are \textit{supervised} methods trained on the train split. * use further additional supervised training samples.
        Cyan: our model is better overall.
        Red: our model is better on specific categories.\footnotemark[2] 
        }
    \label{tab:vqacpv2}
\end{table}
\begin{table}[t]
    \centering
    \small
    \resizebox{\linewidth}{!}{
    \begin{tabular}{@{}lcccc@{}}
        \toprule
        \textbf{Model} & \textbf{All} & \textbf{Yes-No} & \textbf{Num} & \textbf{Others} \\ 
        \toprule
        GVQA~\shortcite{agrawal2018don} & 48.2 & 72.0 & 31.1 & 34.7 \\
        UpDown~\shortcite{Anderson_2018_CVPR} & 65.3 & 81.8 & 44.2 & 56.1 \\
        RUBi~\shortcite{RUBI} & 63.1 & {*} & {*} & {*} \\
        MCAN~\shortcite{yu2019deep} & 70.4 & 85.8 & 53.7 & 60.7 \\
        VilBERT~\shortcite{lu2019vilbert} & 70.5 & {*} & {*} & {*} \\
        LXMERT~\shortcite{tan2019lxmert} & 72.5 & \textbf{88.2} & \textbf{54.2} & \textbf{63.1} \\
        UNITER~\shortcite{chen2019uniter} & \textbf{72.7} & {*} & {*} & {*} \\ \midrule
        ZSL~+~Objects~+~UpDown & 41.4 & 68.1 & 27.6 & 29.4 \\
        ZSL~+~Patches~+~UpDown & 40.6 & 67.8 & 28.4 & 29.2 \\
        ZSL~+~Patches~+~Encoder & \underline{46.8} & \underline{72.1} & \underline{34.4} & \underline{34.1} \\ 
        \midrule
        FSL~+~Objects~+~UpDown & 66.8** & 82.4** & 45.1** & 56.4** \\
        FSL~+~Patches~+~UpDown & 63.4 & 80.2 & 45.2 & 52.1 \\
        FSL~+~Patches~+~Encoder & 65.3 & 80.5 & 48.94 & 56.2 \\ \bottomrule
    \end{tabular}
    }
    \caption[VQA-v2]{VQA-v2 Test-standard accuracies\footnotemark[2]. FSL models are pretrained on synthetic samples, and further finetuned on VQA-v2 train split. * - Scores are not available,  ** - Validation split scores. }
    \label{tab:vqateststd}
\end{table}
\footnotetext[2]{
ZSL refers to zero-shot transfer setting and FSL refers to our models further finetuned on the respective train split. 
\underline{Underline}${\Rightarrow}$unsupervised best, \textbf{bold}${\Rightarrow}$overall best. Baselines are trained on train-split, our models on synthetic data.}
\begin{table}[!h]
    \centering
    \small
    \begin{tabular}{@{}lccc@{}}
        \toprule
        \textbf{Model} & \textbf{All} & \textbf{Binary} & \textbf{Open} \\ 
        \toprule
        CNN~+~LSTM~\shortcite{hudson2018compositional} & 46.6 & 61.9 & 22.7 \\
        UpDown~\shortcite{Anderson_2018_CVPR} & 49.7 & 66.6 & 34.8 \\
        MAC~\shortcite{hudson2018compositional} & 54.1 & 71.2 & 38.9 \\
        BAN~\shortcite{kim2018bilinear} & 57.1 & 76.0 &  40.4 \\
        LXMERT~\shortcite{tan2019lxmert} & \textbf{60.3} & \textbf{77.8} & \textbf{45.0} \\ \midrule
        ZSL~+~Objects~+~UpDown & 30.7 & 50.8 & 17.6 \\
        ZSL~+~Patches~+~UpDown & 31.1 & 52.3 & 16.8 \\
        ZSL~+~Patches~+~Encoder & \underline{33.7} & \underline{55.5} & \underline{21.2} \\ 
        \midrule
        FSL~+~Objects~+~UpDown & 50.4 & 67.5 & 35.1 \\
        FSL~+~Patches~+~UpDown & 46.4 & 64.3 & 31.4 \\
        FSL~+~Patches~+~Encoder & 55.2 & 73.6 & 38.8\\
        \bottomrule
    \end{tabular}
    \caption[GQA]{GQA Validation split accuracies.\footnotemark[2]}
    \label{tab:gqaval}
\end{table}

\section{Results\footnotemark[2]}

\paragraph{Unsupervised Question Answering:} 
Tables \ref{tab:vqacpv2}, \ref{tab:vqateststd} and \ref{tab:gqaval} summarize our results on the three benchmark datasets.
We can observe that our method outperforms specially designed supervised methods for bias removal in VQA-CP; our model with UpDown is $1.1\%$ better than the supervised UpDown.
Under the ZSL setting for VQA-CP, our Encoder model is $6.1\%$ better than UpDown with patches, and $6.5\%$ better than UpDown with Object features, for VQA-v2: $6.2\%$, $5.4\%$ respectively, and for GQA: $2.2\%$, $3.0\%$ respectively.

For VQA-CP, our procedurally generated Q-A pairs and patch-features when used with either UpDown or Encoder are better than the baseline supervised UpDown model, showing the improvements are model-agnostic.
This also shows the merits of using our Q-A generation methods when train and test-sets deviate linguistically.

Most GQA questions require understanding spatial relationships between objects.
Such questions are infrequent in our synthetic training data since captions do not contain detailed spatial relationships among objects. 
Thus, the ZSL performance is not as competitive for GQA when compared to our performance on VQA and VQA-CP. 
Improving spatial and compositional question-answering with weak supervision is an interesting future pursuit.

\paragraph{Fully Supervised Question Answering:}
In the FSL setting, our methods' performance is not far from SOTA methods, even though our method uses significantly fewer annotations (no access to object bounding boxes).
In GQA, the Encoder model performs on par with MAC~\shortcite{hudson2018compositional} and BAN~\shortcite{kim2018bilinear}, which unlike us, use object relationship annotations. 
This suggests that cross-modal transformer layers can learn spatial relations from spatial pyramidal features.

\begin{table}[t]
    \centering
    \small
    \resizebox{\linewidth}{!}{
    \begin{tabular}{@{}llccc@{}}
        \toprule
        & \textbf{Question Generation} & \textbf{VQA-v2} & \textbf{VQA-CP} & \textbf{GQA} \\
        \toprule 
        \multirow{4}{*}{\rotatebox[origin=c]{90}{Updn}} & Template & 26.2 & 25.7 & 11.6 \\
        & Template~+~Para{\&}Back & 28.5 & 27.1 & 14.8 \\
        & QA-SRL & 31.1 & 33.8 & 18.9 \\
        & All & 41.4 & 40.2 & 31.1 \\ 
        \midrule
        \multirow{4}{*}{\rotatebox[origin=c]{90}{Encoder}} & Template & 32.5 & 31.3 & 18.5 \\
        & Template~+~Para{\&}Back & 34.8 & 33.6 & 23.6 \\
        & QA-SRL & 40.3 & 39.8 & 21.4 \\
        & All & \textbf{47.1} &\textbf{ 46.8} & \textbf{33.7} \\ \bottomrule
    \end{tabular}
    }
    \caption{Effect of different pre-training data sources on ZSL Validation split accuracies.}
    \label{tab:qgenablations}
\end{table}


\paragraph{Impact of each question-generation technique:} 
In Table~\ref{tab:qgenablations} we can observe the effect of different question generation techniques. 
All models use spatial image patch features. QA-SRL based questions and the SWA-Loss contribute the most towards gains in performance, and the paraphrased questions provide larger linguistic variation.

\begin{table}[t]
\centering
\small
\begin{tabular}{@{}llccc@{}}
    \toprule
     & \textbf{Patch Resolutions} & \textbf{VQA-v2} & \textbf{VQA-CP} & \textbf{GQA} \\
    \midrule 
    \multirow{5}{*}{\rotatebox[origin=c]{90}{UpDn}}
     & \{1\}            & 18.8 & 19.7 & 11.3 \\
     & \{1, 3\}         & 36.7 & 35.9 & 24.5 \\ 
     & \{1, 3, 5\}      & 40.1 & 39.7 & 29.5 \\ 
     & \{1, 3, 5, 7\}   & \textbf{41.4} & \textbf{40.2} & \textbf{31.1}\\ 
     & \{1, 3, 5, 7, 9\}& 39.8 & 38.4 & 29.3 \\
    \midrule 
    \multirow{5}{*}{\rotatebox[origin=c]{90}{Encoder}}
     & \{1\}            & 26.4 & 27.7 & 15.3 \\
     & \{1, 3\}         & 42.6 & 43.1 & 28.8 \\ 
     & \{1, 3, 5\}      & 44.3 & 45.2 & 30.9 \\ 
     & \{1, 3, 5, 7\}   & \textbf{47.1} & \textbf{46.8} & \textbf{33.7} \\ 
     & \{1, 3, 5, 7, 9\}& 46.2 & 45.4 & 31.2 \\
    \bottomrule
\end{tabular}
\caption{Effect of the number of spatial patches on ZSL performance
\{3,5\} implies division of the image into a 3x3 and 5x5 grid of patches.}
\label{tab:spablations2}
\end{table}

\paragraph{Effect of Spatial Pyramids:}
We study the effect of progressively increasing the number of spatial image patches (i.e., decreasing the patch size).
Table~\ref{tab:spablations2} shows that an optimum exists at grid-size of $7\times7$ after which the addition of smaller patches is detrimental.
Similarly, only using patches of large size does not allow models to focus on specific image regions.
Thus a trade-off exists between global context and region-specific features.
Changing the feature extractor from ResNet-50 to ResNet-101 only results in a minor improvement of $0.01\%$ to $0.30\%$.
Removing visual position embeddings has a significant effect on performance, with a drop of $4.60\%$ to $8.00\%$ in both ZSL and FSL settings.

\paragraph{Impact of Pre-training Tasks:}

\begin{table}[t]
    \centering
    \small
    \begin{tabular}{@{}lcccccc@{}}
        \toprule
        \textbf{Pre-Training Task} & \textbf{VQA-v2} & \textbf{VQA-CP} & \textbf{GQA} \\
        \midrule
        SWA         & 39.1 & 38.3 & 25.4 \\
        MLM+SWA     & 42.4 & 41.5 & 27.8 \\
        MQA+SWA     & 42.0 & 41.2 & 26.6 \\
        MLM+MQA+SWA & 45.6 & 44.9 & 29.7 \\
        MLM+ITM+SWA & 44.7 & 43.6 & 28.9 \\
        \midrule
        \textbf{All} & \textbf{46.2} & \textbf{45.4} & \textbf{31.2}\\
        \bottomrule
        
    \end{tabular}
    \caption{Effect of different pre-training tasks on the ZSL performance for the Encoder model.}
    \label{tab:ptablations}
\end{table}
Table~\ref{tab:ptablations} shows the effect of different pretraining tasks on the downstream zero-shot transfer VQA task. 
We need the SWA task, as it is used to perform the zero-shot QA task. 
The combination of MLM, MQA, and ITM, all of which need image understanding, shows improved performance on the downstream task, indicating better cross-modal representations.

\begin{figure}[t]
    \centering
    \includegraphics[width=0.9\linewidth]{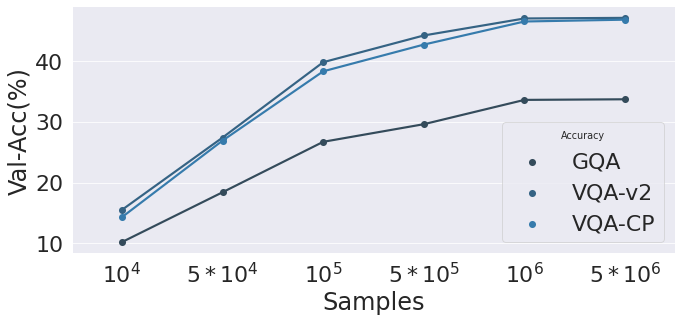}
    \caption{Learning Curve showing validation accuracy vs.\ number of synthetically generated training samples.}
    \label{fig:lr}
\end{figure}    

\paragraph{Effect of size of synthetic training set:} 
Figure~\ref{fig:generalization_types} shows our Encoder model's learning curve for the zero-shot transfer setting trained on our synthetic Q-A pairs. 
The performance stagnates after a critical threshold of $10^6$ samples is reached.
Our experiments also suggest that randomly sampling a set of questions for each image per epoch leads to a $4\%$ gain compared to training on the entire set.
    
\paragraph{Error Analysis:}
Our ZSL method is pretrained on longer phrases and hence tends to generate more detailed answers, such as ``red car" instead of ``car".
Although the SWA loss is designed to encourage a distribution over the shorter phrases, the bias is not entirely removed. 
On automated evaluation, we observe that for $42\%$ of questions, the target answer is a sub-phrase of our predicted answer.
Manual evaluation of 100 such samples shows that $87\%$ of such detailed predicted answers are plausible.
This shows the relevance of learning from captions and quantifies the bias towards short ``true" answers in human-annotated benchmarks, calling for better evaluation metrics that do not penalize VQA systems for producing descriptive or alternative accurate answers.


In the FSL setting, we either finetune our pre-trained QA classifier with the SWA Loss or train a separate feedforward layer from scratch for the task.
The pre-trained QA classifier predicts longer phrases as answers, leading to a drop in accuracy. 
The feedforward layer performs better ($+6\%$), indicating our Encoder captures relevant features necessary to generalize to the benchmark answer-space. 
Note that we do not use object annotations during training, unlike existing methods.

Our error analysis and Figure~\ref{fig:data_tsne_stats} show the shift in question-space and answer-space between synthetic and human-authored Q-A pairs. 
These (along with inadequate evaluation metrics) act as the primary sources explaining the performance-gap between weakly-supervised methods and the fully-supervised setting. 
It remains to be seen whether more sophisticated question generation can be developed to reduce the performance gap further and mitigate the heavy reliance on human annotations.

\section{Discussion and Conclusion}

Prior work~\cite{chen2019uniter,jiang2020defense} has demonstrated that the use of object bounding-boxes and region features leads to significant improvements on downstream tasks such as captioning and VQA.
However, little effort has been dedicated to developing alternative methods that can approach similar performance without relying on dense annotations.
We argue that weakly supervised learning coupled with data synthesis strategies could be the pathway for the V\&L community towards a ``post-dataset era''.\footnote{A. Efros, \textit{Imagining a post-dataset era}, ICML'20 Talk.}
In this work, we take a step towards that goal.
We address the problem of weakly-supervised VQA with a framework for the procedural synthesis of Q-A pairs from captions for training VQA models,  where benchmark datasets can be used only for evaluation.
We use spatial pyramids of patch features to increase the annotation efficiency of our methods.
Our experiments and analyses show the potential of patch-features and procedural data synthesis and reveal problems with existing evaluation metrics.


\section*{Ethical Considerations}
Captions and Question-Answer pairs are both annotated by humans in existing image captioning and visual question answering datasets.
However, captions arguably contain a lesser degree of subjectivity, ambiguity, and linguistic biases than VQA annotations, due to the design of annotation prompts that limit the introduction of these biases.
Our work points to the potential of procedurally generated annotations in providing robustness improvements under changing linguistic priors in VQA test sets (Table~\ref{tab:vqacpv2}).
\citeauthor{hendricks2018women} find that gender bias exists in image-captioning datasets and is \textit{amplified} by models; further research in self-supervised data synthesis could potentially help alleviate such social biases.

\section*{Acknowledgements}
The authors acknowledge support from the DARPA SAIL-ON program W911NF2020006, ONR award N00014-20-1-2332, and NSF grant 1816039, and the anonymous reviewers for their insightful discussion.

\bibliographystyle{acl_natbib}
\bibliography{acl2021}
\section*{Appendix}

\appendix

\section{Synthesized Samples}

Table~\ref{tab:supp_template} shows illustrative examples of Q-A pairs procedurally generated from the image caption using template-based method.
\begin{table}[!h]
    \centering
    \resizebox{\linewidth}{!} {
    \begin{tabular}{@{}p{25mm}>{\raggedright}p{48mm}p{15mm}@{}}
        \toprule
        \textbf{Image}& \textbf{Question} & \textbf{Answer} \\
        \toprule
        \begin{minipage}{25mm}{\includegraphics[width=25mm]{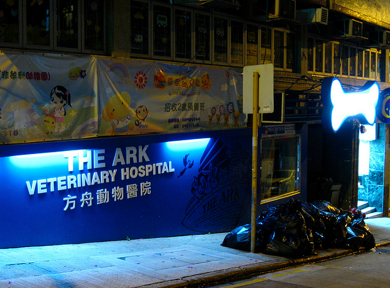}}\end{minipage} & {What are set on the sidewalk outside a veterinary hospital?} & bags \\ \midrule
        \begin{minipage}{25mm}{\includegraphics[width=25mm]{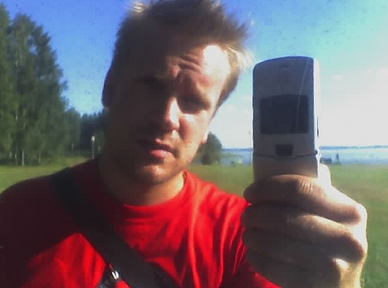}}\end{minipage} & What is the young man holding up in front of his face ? & phone\\ \midrule
        \begin{minipage}{25mm}{\includegraphics[width=25mm,height=20mm]{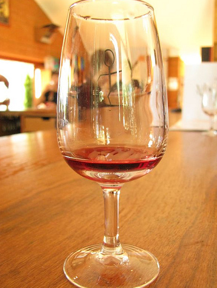}}\end{minipage} & What is almost empty on the table & glass\\\midrule
        \begin{minipage}{25mm}{\includegraphics[width=25mm]{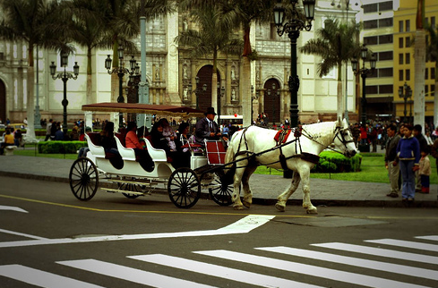}}\end{minipage} & What drawn carriage with passengers in the city & horse \\ \midrule
        \begin{minipage}{25mm}{\includegraphics[width=25mm,height=20mm]{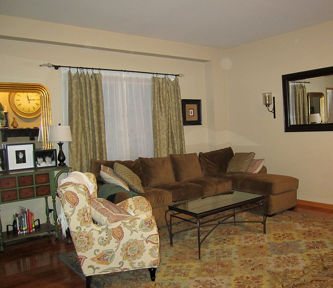}}\end{minipage} & What is the color of the table ? & white\\\midrule
        \begin{minipage}{25mm}{\includegraphics[width=25mm]{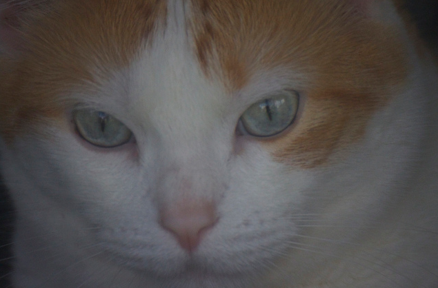}}\end{minipage} & What is the color of the eyes ? & blue \\ \midrule
        \begin{minipage}{25mm}{\includegraphics[width=25mm]{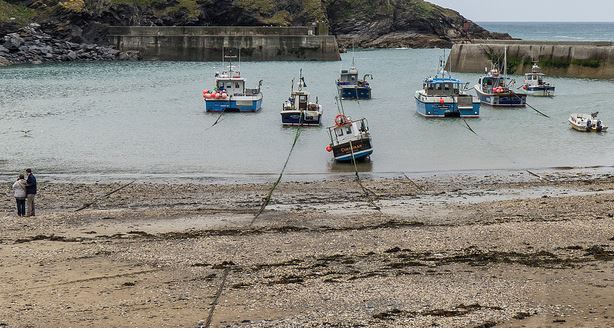}}\end{minipage} & How many boats anchored by ropes close to shore? & 8\\
        \bottomrule
    \end{tabular}
    }
    \caption{Examples of template-based data synthesis}
    \label{tab:supp_template}
\end{table} 
Table~\ref{tab:supp_lol} shows the use of two transformations (T): negation and adversarial words~\cite{gokhale2020vqa} two generate more sentences.
Thus the negation of $Q$ or substitution of a word in $Q$ with an adversarial word results in the new question-answer pair $Q_{new}, A_{new}$.
To increase the linguistic diversity of the questions we use paraphrasing as shown in Table~\ref{tab:supp_paraphrase}.

\begin{table}[!h]
    \centering
    \resizebox{\linewidth}{!} {
    \begin{tabular}{p{1mm}lp{20mm}>{\raggedright}p{22mm}>{\raggedright}p{8mm}>{\raggedright}p{22mm}p{8mm}}
        \toprule
        \textbf{T} & \hphantom & \textbf{Image}& \textbf{Q} & \textbf{A} &  $\mathbf{Q_{new}}$ & $\mathbf{A_{new}}$ \\
        \toprule
        \multirow{10}{*}{\rotatebox[origin=c]{90}{Negation}} && 
        \begin{minipage}{20mm}
            {\includegraphics[width=20mm]{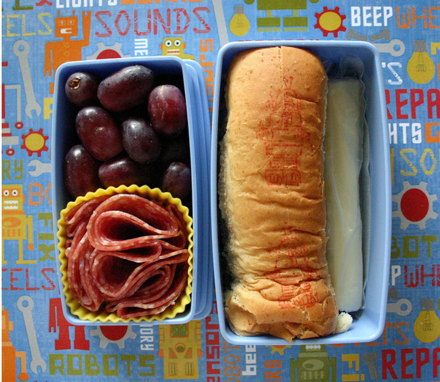}}
        \end{minipage} & Is this bread? & yes & Is this not bread & no \\
        \cmidrule{3-7}
        && \begin{minipage}{20mm}
            {\includegraphics[width=20mm]{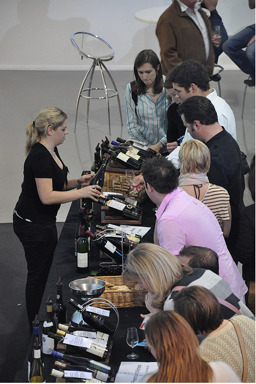}}
        \end{minipage} & What is the color of the woman's shirt? & black & What is not the color of the woman's shirt? & white\\
        \cmidrule{3-7}
        && \begin{minipage}{20mm}
            {\includegraphics[width=20mm]{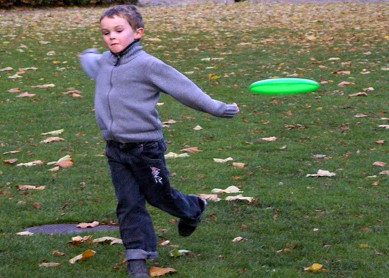}}
        \end{minipage} & Is there a boy? & no & Is there no boy? & yes \\
        \midrule 
        \multirow{9}{*}{\rotatebox[origin=c]{90}{Adversarial}} && 
        \begin{minipage}{20mm}
            {\includegraphics[width=20mm]{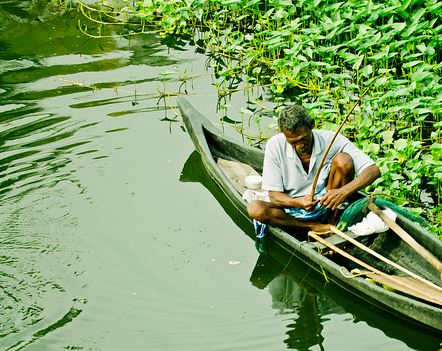}}
        \end{minipage} & Who is sitting in the boat ? & man & Who is sitting in the dining table ? & can't say\\
        \cmidrule{3-7}
        && \begin{minipage}{20mm}
            {\includegraphics[width=20mm]{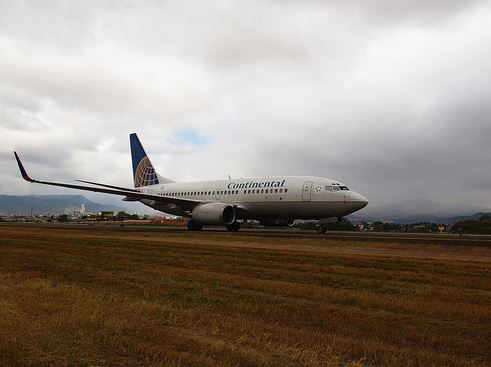}}
        \end{minipage} & How big is the plane ? & large & How big is the car ? & size\\
        \cmidrule{3-7}
        && \begin{minipage}{20mm}
            {\includegraphics[width=20mm]{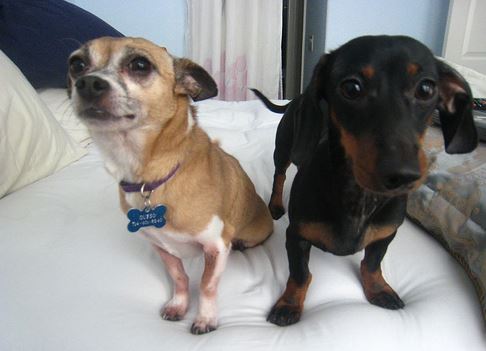}}
        \end{minipage} & How many puppies are on the bed ? & two & How many cats are on the bed? & none\\
        \bottomrule
    \end{tabular}
    }
    \caption{The effect of using transformations (T) to create new Q-A pairs}
    \label{tab:supp_lol}
\end{table}

\section{Dataset Analysis}
In Table~\ref{tab:ans_type}, we compare the distribution per answer-type of our synthetically generated samples with the distribution in the VQA-CP-v2~\cite{agrawal2018don} dataset.
Since we use our synthetic samples as the pre-training data, and do not use VQA-CP samples for training in our zero-shot setup, this comparison displays the shift between the training (synthetic) and test (human annotated VQA-CP) datasets.

\begin{table}[t]
    \centering
    \small
    \begin{tabular}{lcc}
        \toprule
        \textbf{Category} & \textbf{VQA-CP (\%)} & \textbf{Pretraining (\%)}\\
        \midrule
        Yes/No & 41.86 & 50.18 \\
        Number & 11.91 & 8.32 \\
        Other  & 46.23 & 41.45 \\
        \bottomrule
    \end{tabular}
    \caption{Distribution of samples by answer-type in our pre-training dataset and the VQA-CP evaludation dataset.}
    \label{tab:ans_type}
\end{table}

\begin{table}
    \centering
    \small
    \begin{tabular}{lcl}
        \toprule
        \textbf{Hyper-Parameters } & \hphantom & \textbf{Model} \\ 
        \toprule
        Batch Size             && 32-128                         \\ 
        Learning Rate          && ($1e^{-5}$, $5e^{-5}$ )                      \\ 
        Dropout                && 0.1                        \\ 
        Language Layers        && 6                          \\ 
        Cross-Modality Layer   && 4 --- 12                          \\ 
        Optimizer              && BertAdam                   \\ 
        Warmup                 && 0.1                        \\ 
        Max Gradient Norm      && 5.0                        \\ 
        Max Text Length    && 30                         \\ 
        ResNet          && 50 / 101 / 152                       \\
        Epochs          && 10-40 \\
        \bottomrule
    \end{tabular}
    \caption{Hyper-Parameters for our models}
    \label{tab:hyper}
\end{table}

\begin{figure*}
    \centering

    \begin{subfigure}{0.245\linewidth}
        \includegraphics[width=40mm,height=32mm]{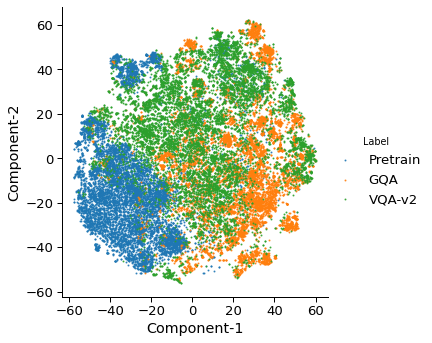}
    \end{subfigure}
    \begin{subfigure}{0.245\linewidth}
         \includegraphics[width=40mm,height=32mm]{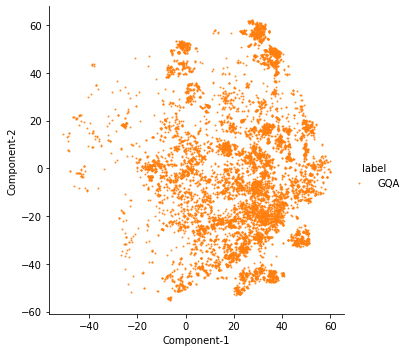}
    \end{subfigure}
    \begin{subfigure}{0.245\linewidth}
         \includegraphics[width=40mm,height=32mm]{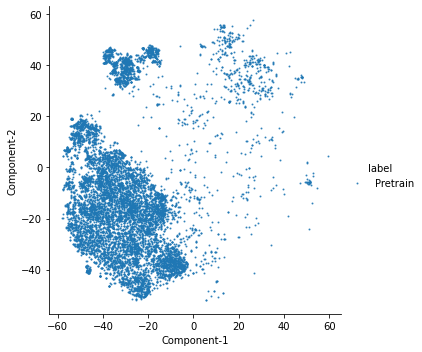}
    \end{subfigure}
    \begin{subfigure}{0.245\linewidth}
         \includegraphics[width=40mm,height=32mm]{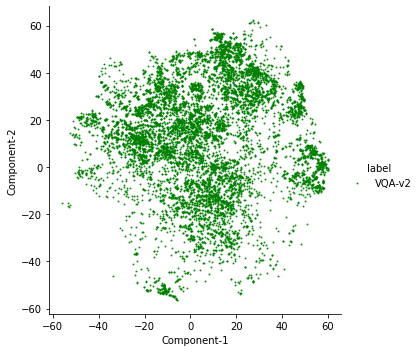}
    \end{subfigure}
    \caption{t-SNE projections of Glove embedding our generated questions, and human-authored VQA-v2 and GQA questions. Blue: our pretraining dataset, Orange: GQA, Green: VQA. L-R: All, GQA, Pretrain, VQA.}
    \label{fig:supp_tsne}    
\end{figure*}

We further analyze this shift, by computing the t-SNE projections of questions using mean-pooled Glove~\cite{pennington2014glove} embeddings for our generated questions and observe the overlap with human-authored questions in VQA and GQA~\cite{hudson2019gqa}.
Figure~\ref{fig:supp_tsne}.
We observe a marked shift between the question clusters for our procedurally generated questions and human annotated questions from VQA and GQA.

\begin{table*}[t]
    \centering
    \resizebox{\linewidth}{!} {
    \begin{tabular}{p{32mm}>{\raggedright}p{53mm}>{\raggedright}p{25mm}>{\raggedright}p{53mm}p{25mm}}
        \toprule
        \textbf{Image}& \textbf{Q} & \textbf{A} &  $\mathbf{Q_{new}}$ & $\mathbf{A_{new}}$ \\
        \toprule
        \multirow{5}{*}{\begin{minipage}{35mm}
            {\includegraphics[width=35mm]{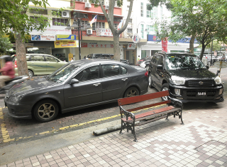}}
        \end{minipage} 
        }
            & How is something parked ? & illegally & How's-what's parked? & illegally\\
         & what does something seem to do ? & park & What do you think something seems to be doing? & park\\
         & Where was parked something? & behind a legally parked car & Do you know where something was parked? & behind a legally parked car \\
         & How many cars are visible ? & 2 & How many cars are we looking at? & 2 \\
         & Is there two cars parked on the sidewalk on the street ? & Yes & There are two cars parked on the sidewalk, right? & Yes \\
        \bottomrule
    \end{tabular}
    }
    \caption{Illustration of using paraphrasing to improve the linguistic variation of our questions and answers.}
    \label{tab:supp_paraphrase}
\end{table*}

\begin{figure*}[!h]
    \centering
    \begin{subfigure}{0.24\linewidth}
         \includegraphics[width=\linewidth]{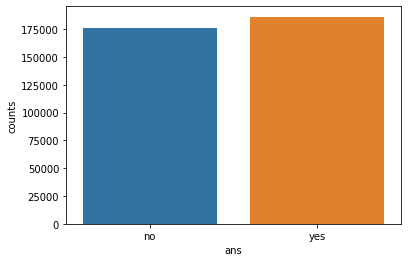}
         \caption{``Yes-No"}
    \end{subfigure}
    \begin{subfigure}{0.2\linewidth}
        \includegraphics[width=\linewidth]{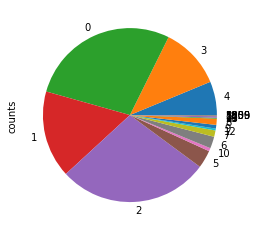}
        \caption{``Numeric"}
    \end{subfigure}
    \begin{subfigure}{0.27\linewidth}
         \includegraphics[width=\linewidth]{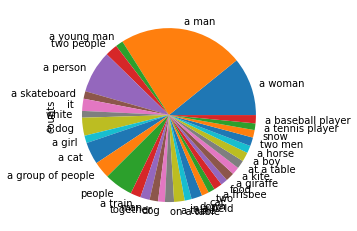}
         \caption{"Other": Highly frequent \\answers (count $>$ 500)}
    \end{subfigure}
    \begin{subfigure}{0.27\linewidth}
         \includegraphics[width=\linewidth]{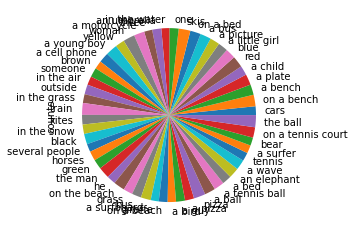}
         \caption{``Other":Answers with \\count between 200 and 500}
    \end{subfigure}
    
    \caption{Distribution of most frequent answers in our Pretraining dataset for each answer-type (yes-no, numeric, and other). Please zoom for details.}
    \label{fig:supp_ans_dist}    
\end{figure*}

Similarly, we also show the distribution of answers in our dataset in Figure~\ref{fig:supp_ans_dist}.
It can be seen that our dataset has a slight imbalance in the proportion of questions with answer ``yes" and ``no".
Numeric answers \textit{0,1,2,3} are most frequent.
Answers about people such as \textit{man, woman, people, person, group of people} are also more common in the dataset.
The remaining answers have a long-tailed distribution, since there are $\sim90k$ unique answers in our dataset compared to $\sim3.5k$ in VQA and $\sim2k$ in GQA.

\section{Training Details}
We use the HuggingFace~\cite{Wolf2019HuggingFacesTS} and PyTorch frameworks \cite{NEURIPS2019_9015}. 
Hyperparameters and other training settings are given in Table~\ref{tab:hyper}.

\end{document}